\pdfoutput=1

\documentclass[11pt]{article}

\usepackage[final]{acl}

\usepackage{times}
\usepackage{latexsym}

\usepackage[T1]{fontenc}

\usepackage[utf8]{inputenc}

\usepackage{microtype}

\usepackage{inconsolata}
\usepackage{graphicx}
\usepackage{array}

\title{Reducing Large Language Model Bias with Emphasis on “Restricted Industries”: Automated Dataset Augmentation and Prejudice Quantification}

\author{Devam Mondal \\ {\bf}
  School of Systems and Enterprises \\
  Stevens Institute of Technology \\
  \texttt{dmondal@stevens.edu} \\\And
  Carlo Lipizzi \\
  School of Systems and Enterprises \\
  Stevens Institute of Technology \\
  \texttt{clipizzi@stevens.edu}\\}

\begin{document}
\maketitle
\begin{abstract}
Despite the growing capabilities of large language models, there exists concerns about the biases they develop. In this paper, we propose a novel, automated mechanism for debiasing through specified dataset augmentation in the lens of bias producers and in the context of “restricted industries” with limited data. We additionally create two new additional metrics, the mb-index and db-index, to quantify bias, considering the idea that bias occurs due to both intrinsic model architecture and dataset quality.
\end{abstract}

\section{Introduction}

In recent years, large language models (LLMs) have revolutionized the field of natural language processing, enabling remarkable advancements in tasks such as text generation, translation, and sentiment analysis. These models, driven by their immense size and pretraining on vast textual corpora, have exhibited impressive capabilities in understanding and generating human-like text. However, beneath the surface of their remarkable achievements lies a profound challenge – the omnipresent issue of bias, defined as the “systematic error that arises as a result of a given choice of the texts used to train language models” \cite{navigli2023a}. Bias in LLMs, often derived from the biases present in the data they are trained on as well as the inherent architecture of the model, has raised ethical concerns and has the potential to reinforce harmful stereotypes and misinformation.

This paper aims to address mechanisms to reduce this bias through a comprehensive approach that utilizes both dataset augmentation and metric creation. More specifically, we aim to mitigate bias in “restricted industries,” more specifically industries where data is limited due to confidentiality or availability of information. Examples of restricted industries include the defense, medical, and financial fields. Because of the documented increase in LLM perplexity and greater embedding of prejudice when trained on small datasets or datasets with biased content, our research takes on a two-fold approach.

Firstly, we explore automated dataset augmentation to mitigate bias, using the concept of a bias producer to describe broad creators of bias, such as ethnicity or sexuality and biasers that serve as specific examples. To measure the actual bias in a text, we use terms (words or phrases) with an intrinsic bias. Some of them can be considered elements of a generic bias - like gender or race - and others are industry or sector specific. 

In this paper, we create a list of those terms and then measure the relevance they have in the training 
dataset. We also define two new metrics for quantifying bias with regards to both datasets and models, the db-index and mb index respectively. These metrics provide a crucial feedback loop for researchers and developers to monitor, analyze, and ultimately minimize bias in LLMs.

\section{Literature Review}

As mentioned previously, in recent years, the proliferation of large language models (LLMs) has revolutionized natural language processing tasks, enabling increasing efficiency in a variety of use cases. However, concerns about biases embedded within these models from the external corpus and human knowledge base have prompted significant research efforts, both to categorize and mitigate biases. The existing literature surrounding this topic can be clustered in 5 groups:

\subsection{Cluster 1: Types and Examples of Bias in the Realm of LLMs}
A large amount of literature has laid down the foundation for categorization of bias in the realm of an LLM environment. Hovy et. al. established 5 sources of bias in natural language processing, asserting that the process of data selection, dataset annotation, input representations (word embeddings), model architecture, and research design can instill prejudice within an LLM \cite{hovy2021a}. Moreover, Navigli et. al. attribute dataset selection and quality as the single greatest “producer” of prejudice, with unbalanced topics and outdated text in corpora, as well as narrow-minded dataset creators instilling bias within LLMs \cite{navigli2023a}. Navigli et. al. also define attributes LLMs exhibit bias against, such as age, culture, nationality and religion, providing examples for each generated by GPT-2, GPT-3, and the BLOOM transformer models \cite{navigli2023a}.

However, the major uncovered topic in this cluster is a method of quantifying the bias. Categorizing the biases and recognizing their source provides a qualitative framework to address them, but does not enable for a quantitative method of treating them. 

\subsection{Cluster 2: Bias in Application of LLMs in Restricted Industries}
There exists a large amount of literature illustrating the various biases of LLMs when applied to “restricted industries.” Here, we define “restricted industries” with data that is unique in nature, as well as confidential. Li et. al. for instance, explores the various biases LLMs exhibit in the medical field, with many of these models being trained on largely English corpora from developed countries, therefore biasing understanding of disease towards high-income nations \cite{li2023a}. Moreover, Mikhailov explores various biases of LLMs in the realm of the military, including the possibilities of offensive hallucinations \cite{mikhailov-a}.

Much like Cluster 1, the challenge in this cluster lies in a lack of a quantitative approach to measure the amount of bias with respect to a restricted industry LLM or dataset. Without this framework, tackling the bias algorithmically is not possible. 

\subsection{Cluster 3: Dataset Bias in the Realm of LLMs}
There also exists literature that provides insight into how datasets that are misrepresentative, poor in quality, or rely on subjective assessments for creation can instill bias within LLMs. Wiegand et. al. demonstrated the issues with the Waseem dataset in regards to the task of detecting abusive language in social media, with the LLMs becoming biased towards sports \cite{wiegand2019a}. As the majority of abusive tweets in the Waseem dataset were disproportionately related to sports, the LLM associated abuse with words such as commentators and announcers. Moreover, the dataset’s tweets were skewed towards 3 authors, with this authorial bias becoming embedded within the LLM. Additionally, Geva et. al. found that datasets reliant on annotators caused LLMs to develop bias, being able to “pick up” annotators that produced large numbers of samples (evidenced by better model performance when annotator id is supplied to the model) \cite{geva2019a}. Yet, these LLMs, reliant on datasets with annotator subjectivity, fail to generalize to new examples created by new annotators. 

The challenge in this cluster is a reliance on human annotators. As mentioned previously, these annotators introduce subjectivity when labeling text, introducing more unintended bias in the process of fine tuning large language models. Novel mechanisms that aim to remediate bias in datasets must therefore do so autonomously, without human intervention.  

\subsection{Cluster 4: Inherent Bias in LLM Architectures}
In addition to the dataset aspect of bias, much literature describes how certain LLM architectures, in particular, the long short-term memory (LSTM) and Transformer, can exhibit bias towards certain characteristics of the human knowledge base. For example, White et. al demonstrated through the creation of artificial languages how LSTMs do not have any preference for word ordering, yet Transformer architectures prefer head-final languages \cite{white2021a}.

The challenge in this cluster is once again quantification. There does not exist a framework to measure this bias. 

\subsection{Addressing and Remediating Bias}
In response, there is a large amount of literature that aims to mitigate bias associated with LLMs. For example, Lee et al. aimed to reduce LLM social bias against certain Korean demographic groups through the creation of KOSBI, a high-quality dataset with contexts (generated through rejection sampling) and sentences generated from the contexts \cite{lee2023a}. Both were then annotated as safe or unsafe. Dixon et al. aimed to reduce bias through mining of additional corpa from an unbiased source (Wikipedia), then created a ‘pinned’  metric to measure fairness based on area-under-curve (AUC) \cite{dixon2018a}. Renaldi et al. explored debiasing through domain adaptation, more specific through finetuning, parameters freezing, and attention matrix training, using metrics like StereoSet (Nadeem et al.) and GLUE (Wang et al.) to measure bias and LLM quality respectively \cite{renaldi-a}\cite{s2021a}\cite{wang2018a}. Gao et al. proposed a distribution alignment loss (DAL) to mitigate bias, first generating biased prompts, then using the DAL to reduce Jensen-Shannon divergence (JSD) between distributions for a masked token when other key parts of the prompt are changed \cite{gao2022a}. Huang et al. suggested eliminating bias through reduction of Wasserstein-1 distance between sentiment distributions of different token collections (each of a different demographic) in a phrase \cite{huang2020a}. This process, named counterfactual evaluation, could be done through either embedding regularization (where cosine similarity between two token collections would be reduced) or sentiment regularization. 

The gaps in this cluster are largely dataset focused. Most approaches aim to correct intrinsic aspects of the LLM itself, and approaches utilizing datasets to debias rely on annotators, which may introduce intrinsic bias. 

Therefore, in this paper, we address the gaps in the applications of Cluster 5 in Cluster 2. More specifically, we propose a novel debiasing mechanism aimed for LLMs in the aforementioned “restricted industries” through automated dataset augmentation. Additionally, we propose a novel quantitative measure of model bias by taking into account performance, as well as dataset bias. 

\section{Approach}
\subsection{Dataset Augmentation}

To reduce bias in LLMs with respect to “restricted industries,” we focus on dataset augmentation approaches defined in Cluster 3. However, due to the unique and confidential nature of the data in these industries, we do not use external, unbiased text. Rather, we use the concept of a bias producer, a “lens” initially containing a set of words known as biasers. Formally, if $\beta$ is a bias producer, at the end of our process, there will be a biaser set {b}, where $b_1$, $b_2$, $b_3$ ... $\in b$ are all examples of $\beta$.

Each entry in the dataset is then swept for members of a bias producer. When the first biaser is met, the entry is recopied, and the biaser is changed with another member of the set. This process repeats $|b| - 1$ times, with all elements of ${b}$ filling in the biaser. Through this mechanism, the dataset broadens in size without reliance on external corpora. 

After this, each entry undergoes content morphism, where each entry is upshifted through contextual word embedding sentence augmentation and downshifted to a short summary to better capture human language. Both are then added to the dataset. 

Note that this method eliminates the need for annotation, a process which can introduce subjectivity and bias. Furthermore, unlike most augmentation processes, such a process is targeted, addressing a single source of bias through phrasing of the bias producer while also addressing sentiment. 

\subsection{LLM Bias Classification}
To assess the performance of models after being fine-tuned on an augmented dataset, we propose a new metric called the mb-index. This metric provides an “index of bias with respect to performance per data entry trained on.” Formally, given a dataset ${d}$, perplexity $p(d)$, stereotype score $s(d)$, mb-index is defined as:

$$\frac{p(d) * s(d)}{|d|}$$

The stereotype score, a new metric, is a result derived from an extension of the Intersentence Context Association Test Nadeem et. al proposed in conjunction to the StereoSet score \cite{s2021a} However, rather than the LLM “picking” the best answer to the context provided in a multiple-choice setting, it generates a 30-character continuation of the context, defined as $I$.  

Given three choices, one reinforcing a stereotype ($A$), the other reinforcing the anti-stereotype ($B$), and the third being a nonsensical sentence ($C$), the cosine similarity between I and each option is calculated. The greatest similarity is then used to classify the generated text as stereotypical, anti-stereotypical, or nonsensical. This process is continued through each entry of the StereoSet dataset. Note that this process can be extrapolate to any dataset that contains stereotypical and anti-stereotypical entries.

\begin{figure*}

 \center

  \includegraphics[width=\textwidth]{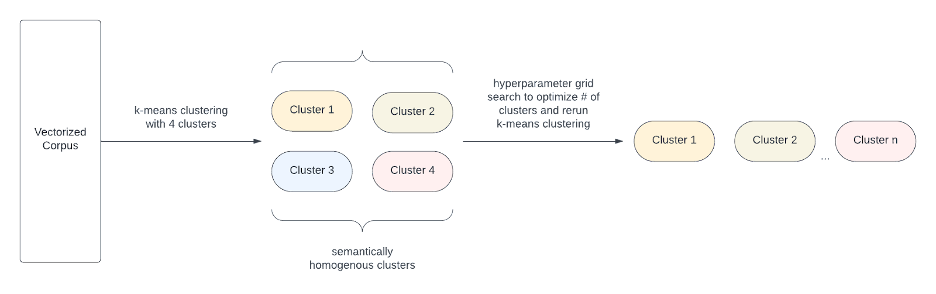}

  \caption{Obtaining semantically homogenous clusters through k-means clustering and hyperparameter grid search.}

  \label{AAA}

\end{figure*}

From this definition, we assert that the stereotype score is the proportion of continuations classified as stereotypical with respect to all continuations not marked as nonsensical:

$$\frac{I_A}{I_B + I_C}$$

For an ideal LLM, the stereotype score should be near 0, as the model’s responses should not be similar to stereotypical responses. It is fine if the model’s responses are similar to the anti-stereotypical responses, as the training procedure promotes the model to “think anti stereotypically,” where a variety of different qualities can be attributed to a lens. 

A “good” mb-index is one that is as close to 0, where stereotype score is ideally and perplexity is ideally minimized to 0. Note that to compare mb-index, two LLMs must use the same reference dataset.

\subsection{Dataset Bias Classification}

However, bias is not just limited to the model, but also a given dataset. Thus, we propose another metric called the db-index. This metric quantifies the amount of bias present in a dataset with regards to cosine similarity. Formally, cosine similarity is a numerical value that shows how “close” two vectors are in a space. Given vector $A$ and $B$, it is defined as:

$$\frac{A \cdot B}{||A||||B||}$$

Given a target dataset $d_t$ and a comparison dataset $d_c$ (containing biased and abusive language), a random entry $e_c \in d_c$ is picked. Then, cosine similarity, $d_{\cos \theta}$ between the vector of the comparison entry and each entry $e_d \in d_t$ is calculated and summed:

$$d_{\cos \theta} = \sum_{i = 1}^{|d_t|}\frac{e_{d,i} \cdot e_{c}}{||e_{d,i}|||e_{c}||}$$

To obtain a dataset’s db-index, we first convert each entry in the corpus into an embedding vector. We then segment the corpus into semantically homogeneous groups using k-means clustering. Because the initial number of clusters is difficult to set, we run clustering with an arbitrary value of 4 clusters, then conduct hyper-parameter tuning through grid search to determine the optimal number of clusters. Then, k-means clustering is conducted again. Figure 1 above shows this process.

The above algorithm is then used on each cluster $c$ (each entry of the cluster comprising the target dataset) created, yielding $d_{\cos \theta}$. $d_{\cos \theta}$ is then divided by the cluster’s size $|c|$ to yield $d_{bc}$, average similarity to an offender over each entry: 

$$db_c = \frac{db_{\cos \theta}}{|d|}$$

The total $db$ is then found by averaging all of the clusters’ db-index:
	
$$db=\frac{db_c}{k}$$

Therefore, with regularization in respect to dataset length, a “good” db-index is one close to 0 (each entry is not very similar to the offender, averaged out). 

\begin{figure*}[h]

 \center

  \includegraphics[width=\textwidth]{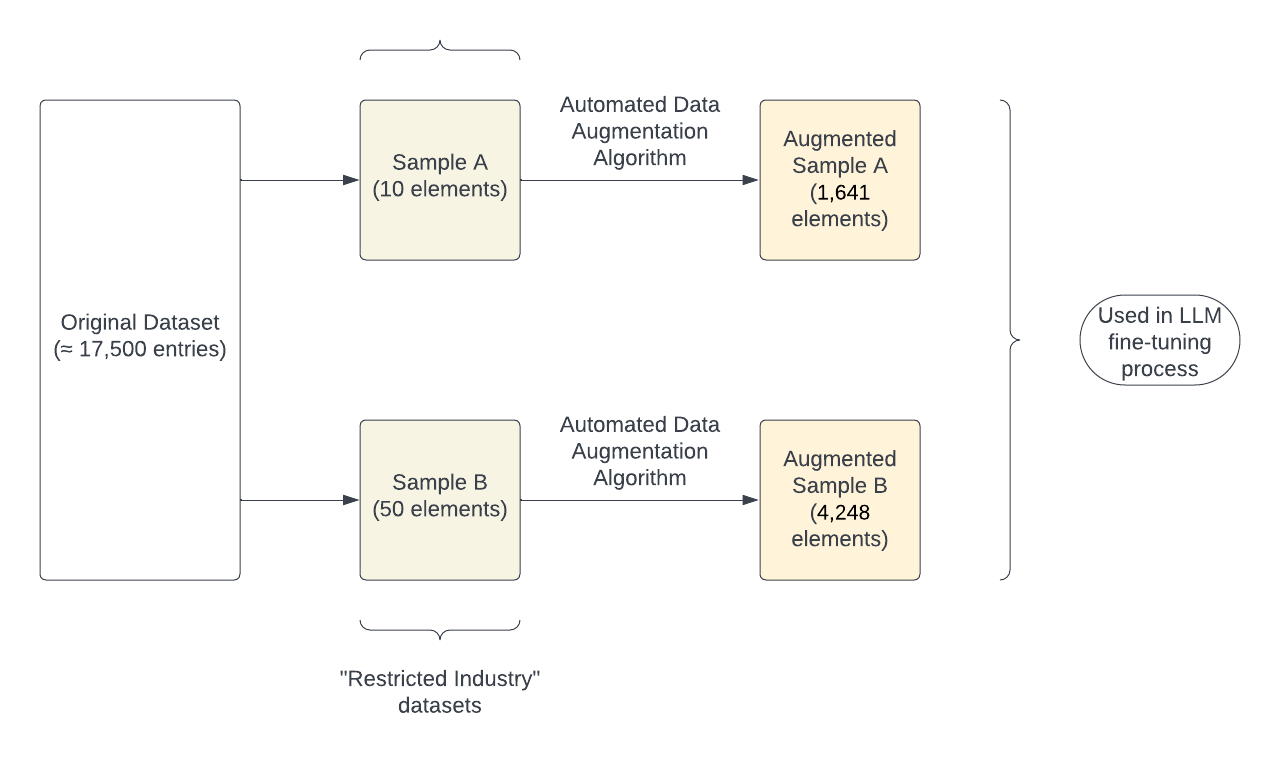}

  \caption{Obtaining semantically homogenous clusters through k-means clustering and hyperparameter grid search.}

  \label{AAA}

\end{figure*}

We define datasets as biased if their db-index exceeds 0.5. Note that to compare the bias of two datasets, they must utilize the same comparison dataset, and the comparison dataset must contain entries that are similar in syntactical structure and bias magnitude.

\section{Method}
We first began with the dataset augmentation procedure. Given the focus of the paper is on “restricted industries,” we sought to augment a dataset of government reports with regards to military and other classified content. This dataset initially contains around 17,500 entries. 

To imitate a situation with a lack of available data, our dataset augmentation method focused on two small subsets of the dataset: Sample A, containing 10 elements, and Sample B, containing 50 elements.

We then conducted dataset augmentation with the lens of ethnicity, the biaser set containing twenty different races generated through a LangChain process with ChatGPT3.5 content generation. The above process, detailed in Part 1 of Approach, was used twice. First, Sample A was augmented to produce a dataset of size 1,641 elements. Secondly, Sample B was augmented to produce a dataset of size 4,248 elements. Figure 2 provides a flow chart of this entire process.

We then calculated the db-index of both Sample A and Sample B, as well as their augmented counterparts, with regards to implicit bias. Table 1 shows the results.

It is important to note that the db-index calculated here was based off a comparison dataset focused on measuring implicit bias. Different aspects of bias can be measured by changing the comparison dataset.

 \begin{table}[!ht]
     \centering
     \begin{tabular}{|c|c|}
          \hline
                \textbf{Dataset} & \textbf{db-index}\\
               \hline
               Sample A & 0.56\\
               \hline
               Sample A (augmented) & 0.49\\
               \hline
               Sample B & 0.71\\
               \hline
               Sample B (augmented) & 0.65\\
               \hline
               \end{tabular}
               \caption{The db-indices of the four datasets.}
 \end{table}
         
Next, four LLMs were chosen to be fine-tuned on the augmented data and original data. All four LLMs were Meta AI’s LLaMa 13b Chat models with HuggingFace formatted weights and biases. Each LLM was fine-tuned on different data with the following:

\begin{table*}[h]
\begin{center}
\begin{tabular}{|l|l|l|l|}
\hline
LLM & Perplexity & Stereotype Score & mb-index                          \\ \hline
A   & 6.4660     & 0.55             & 2.16 x $10^{-3}$ \\ \hline
B   & 6.2920     & 0.52             & 7.65 x $10^{-4}$ \\ \hline
C   & 4.9290     & 0.45             & 1.36 x $10^{-3}$ \\ \hline
D   & 4.9290     & 0.45             & 5.24 x $10^{-4}$ \\ \hline
\end{tabular}
\caption{Performance and bias metrics for the four LLMs}
\end{center}
\end{table*}

\begin{itemize}
  \item LLM A was fine-tuned on a subsection of the original dataset containing the same number of samples as augmented Sample A.
  \item LLM B was fine-tuned on a subsection of the original dataset containing the same number of samples as augmented Sample B.
  \item LLM C was fine-tuned on augmented Sample A. 
  \item LLM D was fine-tuned on augmented Sample B. 
  
\end{itemize}
                  
All LLMs were fine-tuned in a causal language modeling process and through QLoRA (Quantized Low-Rank Adaptation). 

After fine-tuning, each LLM’s perplexity, Stereotype score, and mb-index was calculated using the original dataset as the reference. Table 2 shows the results.

\section{Results and Discussion}

As seen in Table 1, the automated dataset algorithm is able to reduce the db-index of a dataset. The augmented datasets had a substantial decrease in db-index compared to their original counterparts.

As seen in Table 2, LLMs C and D, fine-tuned on the augmented datasets, have less perplexity compared to LLMs A and B. This suggests that augmented datasets, created through the algorithm mentioned above, can increase LLM performance.

Additionally, the stereotype scores for LLMs C and D are also less compared to LLMs A and B, suggesting that the dataset augmentation mechanism reliant on a bias producer “lens” that substitutes in members of a biaser set is effective at removing LLM tendency towards stereotypical responses.

Therefore, because LLMs C and D have lower perplexities and stereotype scores, they, in the quantitative measures described above, are less biased due to a lower mb-index.

\section{Limitations and Future Research Directions}
The datasets used to calculate db-index on were in the scale of tens of thousands due to limits on public data available. It would be beneficial to see db-indices being produced for datasets containing millions of records in order to assess the efficiency of our algorithm. 

Additionally, the LLMs fine-tuned were medium-sized (13 billion parameters). It would be beneficial to see larger LLMs (70 billion parameters or more) being fine-tuned on datasets augmented through our bias-reducing approach and their mb-index performance.

\section{Conclusion}
A pressing matter in the ever-evolving field of natural language processing is the bias present in large language models. In this paper, we lay out a mechanism to tackle bias caused by training and fine-tuning data within large language models through an automated augmentation algorithm based on bias producers. We also provide ways to quantify bias inherent to datasets and large language models through the db-index and mb-index accordingly.

We hope to continue democratizing our work in this paper by creating an online platform where natural language processing enthusiasts and professionals can see the bias within their large language models and datasets before implementing them in their systems.

\bibliography{custom.bib}

\end{document}